\documentclass[sigplan,screen]{acmart}

\AtBeginDocument{%
  \providecommand\BibTeX{{%
    \normalfont B\kern-0.5em{\scshape i\kern-0.25em b}\kern-0.8em\TeX}}}

\copyrightyear{2024}
\acmYear{2024}
\setcopyright{acmlicensed}\acmConference[ASPLOS '24]{29th ACM International Conference on Architectural Support for Programming Languages and Operating Systems, Volume 2}{April 27-May 1, 2024}{La Jolla, CA, USA}
\acmBooktitle{29th ACM International Conference on Architectural Support for Programming Languages and Operating Systems, Volume 2 (ASPLOS '24), April 27-May 1, 2024, La Jolla, CA, USA}
\acmDOI{10.1145/3620665.3640426}
\acmISBN{979-8-4007-0385-0/24/04}

\usepackage{amsmath,amsfonts} 
\usepackage{mathtools}
\newcommand{\pluseq}{\mathrel{+}=}
\usepackage{xspace}

\usepackage{algorithmic}
\usepackage{algorithm}
\algsetup{indent=0.5em}

\usepackage{graphicx}
\usepackage{textcomp}
\usepackage{multirow}

\usepackage{fancyhdr}

\usepackage{pifont} 

\usepackage{hyperref}

\def\BibTeX{{\rm B\kern-.05em{\sc i\kern-.025em b}\kern-.08em
    T\kern-.1667em\lower.7ex\hbox{E}\kern-.125emX}}

\usepackage{array}

\usepackage{xcolor}
\definecolor{commentgreen}{rgb}{0,0.6,0}

\newcommand{\mycomment}[1]{\textcolor{commentgreen}{\texttt{#1}}}

\newcommand{\ourframework}{MaxK-GNN\xspace} 

\newcommand{\takeaway}[1]{\noindent\rule{\columnwidth}{1.0pt}
\textbf{Key takeaway}: \textit{{#1}}\newline
\noindent\rule{\columnwidth}{1.0pt}}

\usepackage{tikz}

\newcommand{\blackcircled}[1]{
  \begin{tikzpicture}[baseline={(char.base)}]
    \node[draw, circle, fill=black, text=white, inner sep=1pt] (char) {#1};
  \end{tikzpicture}
}

\usepackage{xcolor}
\usepackage{soul}

\newcommand{\revA}[1]{\sethlcolor{white}\hl{#1}}
\newcommand{\revB}[1]{\sethlcolor{white}\hl{#1}}
\newcommand{\revC}[1]{\sethlcolor{white}\hl{#1}}
\newcommand{\revD}[1]{\sethlcolor{white}\hl{#1}}

\begin{document}

 \title{\ourframework: Extremely Fast GPU Kernel Design for Accelerating Graph Neural Networks Training}

\author{Hongwu Peng$^{1,*}$, Xi Xie$^{1,*}$, Kaustubh Shivdikar$^2$, MD Amit Hasan$^1$, Jiahui Zhao$^1$, Shaoyi Huang$^1$, Omer Khan$^1$, David Kaeli$^2$, Caiwen Ding$^1$}
\affiliation{$^*$These authors contributed equally. \country{}}
\affiliation{\fontsize{11pt}{11pt}\selectfont \institution{$^1$University of Connecticut \country{USA}. 
 $^2$ Northeastern University \country{USA}.}
 \country{}}
\affiliation{\fontsize{9pt}{9pt}\selectfont
$^{1}$\{hongwu.peng, xi.xie, amit.hasan, jiahui.zhao, shaoyi.huang, khan, caiwen.ding\}@uconn.edu \\
$^{2}$\{shivdikar.k, d.kaeli\}@northeastern.edu
\country{}
}

\renewcommand{\shortauthors}{Peng and Xie, et al.}

\begin{abstract}

In the acceleration of deep neural network training, the graphics processing unit (GPU) has become the mainstream platform. GPUs face substantial challenges on Graph Neural Networks (GNNs), such as workload imbalance and memory access irregularities, leading to underutilized hardware. Existing solutions such as PyG, DGL with cuSPARSE, and GNNAdvisor frameworks partially address these challenges. However, the memory traffic involved with Sparse-Dense Matrix Matrix Multiplication (SpMM) is still significant.

We argue that drastic performance improvements can only be achieved by the vertical optimization of algorithm and system innovations, rather than treating the speedup optimization as an "after-thought" (i.e., \textit{(i) given a GNN algorithm, designing an accelerator, or (ii) given hardware, mainly optimizing the GNN algorithm}). In this paper, we present \ourframework, an advanced high-performance GPU
training system integrating algorithm and system innovation. (i) We introduce the MaxK nonlinearity and provide a theoretical analysis of MaxK nonlinearity as a universal approximator, and present the Compressed Balanced Sparse Row (CBSR) format, designed to store the data and index of the feature matrix after nonlinearity; (ii) We design a coalescing enhanced forward computation with row-wise product-based Sparse Matrix-Matrix Multiplication (SpGEMM) Kernel using CBSR for input feature matrix fetching and strategic placement of a sparse output accumulation buffer in shared memory; 
(iii) We develop an optimized backward computation with outer product-based and Sampled Sparse Matrix Dense Matrix Multiplication (SSpMM)  Kernel.

We conduct extensive evaluations of \ourframework and report the  system  training time. Experiments show that \ourframework system could approach the  speedup limit according to Amdahl's law. We achieve comparable accuracy to SOTA GNNs, but at a significantly increased speed: 3.22$\times$/4.24$\times$ speedup (vs. 5.52$\times$/7.27$\times$) on Reddit compared to DGL and GNNAdvisor implementations. Our implementation can be found on GitHub\footnote{\url{https://github.com/harveyp123/MaxK-GNN}}.

\end{abstract}



\maketitle 

\vspace{-1mm}
\section{Introduction}

Graph Convolutional Networks (GCNs), a specific type of Graph Neural Networks (GNNs), have garnered significant attention in recent years due to their unparalleled capability to extract latent information from graph data~\cite{hu2020open, kipf2016semi, liu2020graphsage}. The field of GCNs manifests in a myriad of important practical applications, including the prediction of cascading power-grid failures~\cite{liu2020guiding}, traffic forecasting~\cite{jiang2022graph}, recommendation systems~\cite{wu2020graph, ying2018graph}, and drug discovery~\cite{bongini2021molecular}. In the design and acceleration of GNN training, GPU platforms have become the prevalent choice. Conventional GCN acceleration processes a graph feature matrix ($X$) by multiplying it with a dense small weight matrix ($W$), followed by multiplying the resultant output with a highly sparse irregular adjacency matrix ($A$) via Sparse-Dense Matrix Matrix Multiplication (SpMM)~\cite{dgl_amazon}.

\begin{figure}[t] \centering
    \includegraphics[ width =0.99 \linewidth]{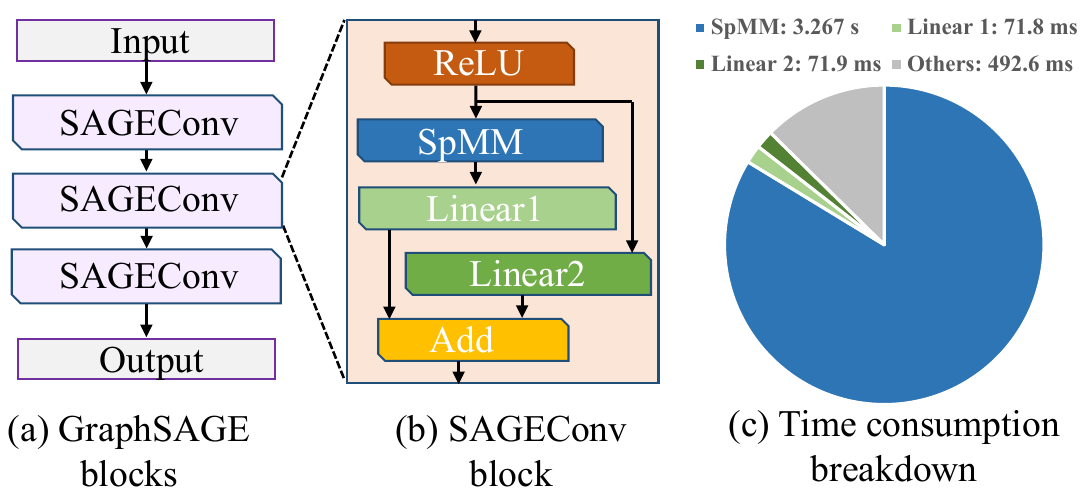}
    \vspace{-2mm}
     \caption{
GraphSAGE structure analysis: latency breakdown of full-batch GraphSAGE training on the \textit{ogbn-proteins} dataset over 30 epochs, with 256 hidden dimensions. GPU platform: Nvidia A100. 
     }
    \label{fig:breakdown} 
\vspace{-5mm}
\end{figure}

Addressing the demands for high-performance and efficient GNN systems has led to two primary research trends: algorithmic optimization and hardware-level enhancement. The former encapsulates methods such as graph reordering, e.g., GNNAdvisor~\cite{wang2021gnnadvisor}, run-time community detection, e.g., I-GCN~\cite{geng2021gcn}, and graph partitioning, e.g., GCoD~\cite{you2022gcod}. Conversely, hardware-level approaches focus on workload balancing and efficient hardware mapping, with specific work tackling workload imbalance stemming from irregular input data with power-law distributed non-zero elements, {e.g., AWB-GCN \cite{geng2020awb}, FlowGNN \cite{sarkar2022flowgnn}, MergePath-SpMM \cite{shan2023mergepath}, GROW \cite{hwang2023grow}, G-CoS \cite{zhang2021g}, ENGN \cite{liang2020engn}}~\cite{geng2020awb, shan2023mergepath, zhang2021g, abadal2021computing, liang2020engn, liang2020deepburning, auten2020hardware, hwang2023grow, sarkar2022flowgnn}.

\textbf{Challenges.} Despite the advancements, there are grant challenges. 
Many existing accelerators,  including AWB-GCN \cite{geng2020awb} and GCoD~\cite{you2022gcod}, are FPGA based~\cite{geng2020awb, geng2021gcn, you2022gcod} or ASIC based \cite{yan2020hygcn, byun2020cryocore} which are typically not open-sourced, and are user-unfriendly. They require specialized hardware such as on-chip distribution networks and comprehensive graph preprocessing support to address the workload imbalance caused by SpMM (referred to as "evil rows")~\cite{geng2020awb}. In comparison,
existing GPU systems provide open-sourced and user-friendly implementations,
however, 
\revD{they are} still far from meeting
performance limits.  Using the profiling results of full batch GraphSAGE~\cite{liu2020graphsage} training as an illustration, shown in Fig.~\ref{fig:breakdown}. The computation and memory demands associated with the SpMM kernel are the major bottlenecks during the training process, contributing to over $83.6\%$ of the total training time. 
More specifically, the GPU's multi-level memory hierarchy~\cite{jia2018dissecting} and SpMM's usage of memory-efficient formats (e.g., compressed sparse row (CSR))
create difficulties in shared memory buffering design and hinder the exploitation of memory locality~\cite{shivdikar2021smash}. 

\textbf{Research Gap.} We summarize the root causes of the above inefficiencies as:
(i) \textit{Memory Traffic Challenges in GPU-based Frameworks:} 
Existing works adopt a row-wise multiplication approach which employs nonzero-grouping techniques, e.g., GNNAdvisor~\cite{wang2021gnnadvisor}, thereby transferring atomic accumulation into shared memory which resides in a streaming multiprocessor (SM). Although this approach mitigates the cost of atomic accumulation in global memory, it still requires a substantial number of global memory transactions to access the input feature matrix, resulting in total memory traffic scaling linearly with the hidden dimension and number of nonzeros~\cite{baruah2021gnnmark}.
The linear scaling with \revC{original hidden dimension $dim_{origin}$} and $nnz$ exacerbates this problem. MergePath~\cite{shan2023mergepath} further resolves SpMM workload imbalance issues using a binary search-based warp mapping, but is less effective when the hidden dimension is large. 
(ii) \textit{Algorithmic Limitations and Resource Waste:} Prevailing algorithmic methods, such as graph partitioning~\cite{wan2022bns} and graph sampling~\cite{yang2023betty}, which are tailored to address large-scale graph training challenges, frequently lead to a reduction in accuracy~\cite{wan2022bns} and accrue overhead in communication~\cite{wan2022bns} and subgraph sampling, as well as redundant computing~\cite{yang2023betty}. This architecture-oblivious workflow consistently results in inefficient hardware utilization. 
These gaps underline the pressing need for \textbf{sustainable} acceleration solutions.

\begin{figure}[t]\centering
    \includegraphics[
    clip, 
    width =0.99 \linewidth]{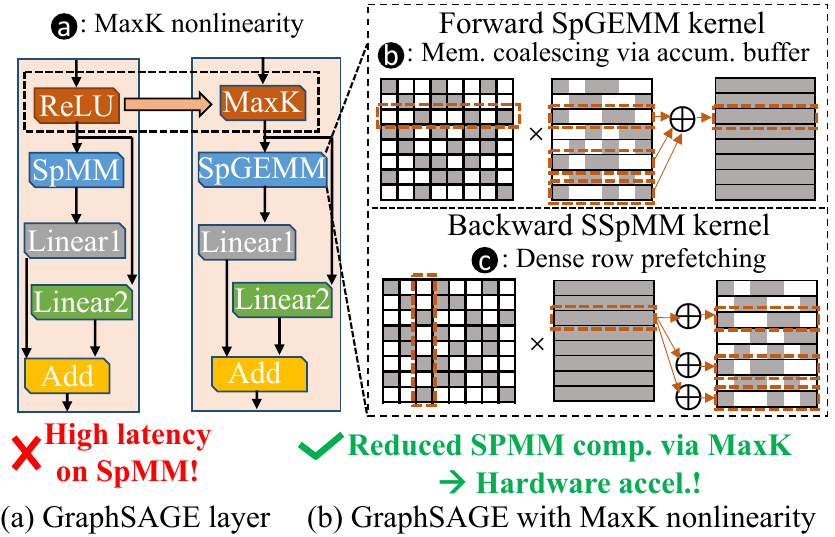} 
    \vspace{-2mm}
     \caption{GraphSAGE layer example with (a) ReLU (b) MaxK nonlinearity. SSpMM: sampled sparse matrix dense matrix multiplication. }
    \label{fig:contribution} 
\vspace{-3mm}
\end{figure}

\textbf{Proposed Research.}
We argue that drastic performance improvements can only be achieved by the vertical integration and optimization of algorithms and system innovations. Our approach is substantially different. Rather than treating the sustainability optimization as an "after-thought" (i.e., \textit{(i) given a GNN algorithm, designing an accelerator, or (ii) given a platform, primarily optimizing the GNN algorithm}), we propose a set of GNN paradigms that work cooperatively at both the algorithm and GPU system levels to deliver strong performance scaling. Our target is a high accuracy, high performance, and low latency GNN training system.

In this work, we introduce \ourframework, an advanced GPU training system 
integrating algorithm and system innovations.
Our design 
significantly outperforms
the state-of-the-art (SOTA) GPU-based GNN training solutions, including GNNAdvisor~\cite{wang2021gnnadvisor} and DGL~\cite{dgl_amazon}. 
\ourframework is strategically constructed on the PyTorch framework~\cite{paszke2019pytorch} for its front-end, and further extends the GPU's computational capabilities by customizing the MaxK nonlinearity to select the top-$kth$ element for each node embedding and implementing innovative Sparse Matrix-Matrix Multiplication (SpGEMM) and Sampled Sparse Matrix Dense Matrix Multiplication (SSpMM) kernels using C++/CUDA. 

The design of \ourframework system is focused on three core \textbf{contributions},
also illustrated in Fig.~\ref{fig:contribution}:

\blackcircled{a} \textit{Node-Balanced Feature Dimension Reduction through MaxK Nonlinearity:} We introduce the MaxK nonlinearity, and provide a theoretical analysis of MaxK nonlinearity as a universal approximator. We present the Compressed Balanced Sparse Row (CBSR) format, designed to store the data and index of the feature matrix after nonlinearity. This approach not only facilitates memory coalescing, but also significantly reduces traffic on the GPU platform. Experiments show that we can reduce the effective feature map dimension from 256 to 16 with a minor accuracy drop.

\blackcircled{b} \textit{Coalescing Enhanced Forward Computation with Row-wise Product-Based SpGEMM Kernel:} This component encompasses: (i) the utilization of the CBSR format for right-hand matrix fetching, leading to a notable memory traffic reduction. For example, Reddit dataset with the original hidden dimension as 256 and \revC{MaxK $k$ value} as 16, can reduce the global memory traffic by 90.6\% compared to SpMM. 
(ii) the strategic placement of a sparse output accumulation buffer in shared memory, enabling coalesced global memory accumulation on the output matrix, while maintaining the same accumulation efficiency as a conventional SpMM design. 

\blackcircled{c} \textit{Optimized Backward Computation with Outer Product-Based SSpMM Kernel Design:} This segment focuses on the acceleration of the computation pattern (sparse $\times$ dense = sparse). Leveraging a dense row prefetching technique, we effectively transfer irregular memory accesses from global memory to shared memory. The subsequent irregular shared memory fetching is facilitated using the CBSR index, followed by atomic accumulation of the CBSR data in global memory. The proposed SSpMM design ensures coalesced memory transactions across all stages, substantially reducing global memory consumption by more than 90\% (Reddit dataset with original hidden dimension as 256 and $k$ as 16).

We conduct extensive evaluations of \ourframework system within the context of a single-GPU, full-batch, GNN training workload. We report the system  training time rather than floating point operations per second (FLOPS) analysis. Experiments show that our \ourframework system could approach the  speedup limit according to Amdahl's law~\cite{gustafson1988reevaluating}. The performance gaps between our results and the limits, e.g., 3.22$\times$/4.24$\times$ compared to 5.52$\times$/7.27$\times$ for Reddit dataset using \ourframework with GraphSAGE, is from the accumulation stage of SpGEMM and dense row prefetching stage of SSpMM, which empirically are difficult to further optimize. 

The introduced MaxK non-linearity and kernel design are not confined to the specific framework, but exhibit compatible with other SOTA GNN training systems, including PyG~\cite{fey2019fast} and DGL~\cite{dgl_amazon}. Furthermore, the adaptability of these novel constructs aligns with current methods employed in graph partitioning~\cite{chiang2019cluster, wan2022bns} and graph sampling~\cite{yang2023betty, zeng2019graphsaint}.

\section{Background and Related work}

\subsection{Graph Convolution Network}
\begin{figure}[t]
\centering
    \includegraphics[
    clip, 
    width =0.99 \linewidth]{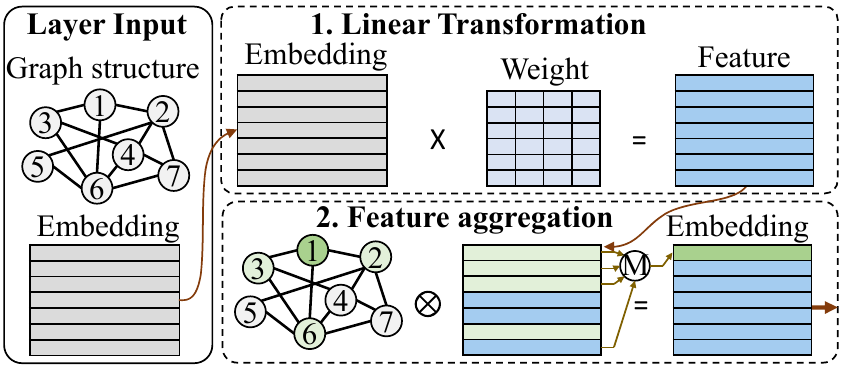} 
    \vspace{-2mm}
     \caption{
     Computational workflow of a GCNConv layer.}
    \label{fig:GCN_structure} 
\vspace{-4mm}
\end{figure}

Graph Convolutional Networks (GCNs)~\cite{kipf2016semi} are stacks of GCNConv layers. An example of a GCNConv layer is shown in Fig.~\ref{fig:GCN_structure}. 
We define a graph $G=(\mathcal{V}, \mathcal{E}, A)$ which contains $|\mathcal{V}|$ nodes and $|\mathcal{E}|$ edges. 
The adjacency matrix $A$ has the shape of $(|\mathcal{V}| \times |\mathcal{V}|)$, usually with high sparsity. 
Each non-zero entry $A(i,j)$ corresponds to an edge between $i$ and $j$. 
Each node is associated with an $\mathcal{F}$-dimensional feature embedding vector, and $X \in \mathbb{R}^{\mathcal{|V|} \times \mathcal{F}}$ represents the feature embedding matrix for all nodes. 
The forward propagation of the $l$-th GCNConv layer can be split into 2 stages: (1) linear transformation $Y^l = X^l W^l \label{linear-transformation}$ and (2) feature aggregation $X^{l+1} = \sigma(A' Y^l) \label{feature-aggregation}$. 
Where $X^l \in \mathbb{R}^{|\mathcal{V}| \times \mathcal{F}_l}$ is the feature embedding matrix at the $l$-th layer for all nodes, $W^l \in \mathbb{R}^{\mathcal{F}_l \times \mathcal{F}_{l+1}}$ is the weight matrix for linear transformation which will be learned during the GCN training. 
The feature aggregation stage calculates the feature embedding matrix for the next layer, where $A' \in \mathbb{R}^{|\mathcal{V}| \times |\mathcal{V}|}$ is the normalized and regularized adjacency matrix, $\sigma$ is the activation function, typically element-wise ReLU.
Different varients of GCNs, such as GraphSage~\cite{hamilton2017inductive} and Graph Isomorphism Network (GIN)~\cite{hou2019measuring}, use similar structure and can reuse the same forward propagation abstraction as GCNs. 

\subsection{GNN Acceleration}
The PyTorch Geometric (PyG) software stack~\cite{fey2019fast} and similar proposals like HP-GNN~\cite{lin2022hp}, LL-GNN~\cite{zhang2022low}, and FlowGNN~\cite{sarkar2022flowgnn} utilize message-passing primitives such as \textit{scatter} and \textit{reduce} for GNN training on GPUs and FPGA overlays. 
These primitives incur substantial memory and storage overheads, leading to inefficiency and poor memory bandwidth utilization. None of these approaches effectively enhance workload balance or address data locality issues within GNN training and inference.

Several existing open-source GPU and FPGA acceleration frameworks are aimed at enhancing GNNs. GNNAdvisor~\cite{wang2021gnnadvisor} uses warp-level partitioning for distributed neighborhood workload but may cause load imbalance and its kernel performance, mainly improved by the Rabbit order~\cite{arai2016rabbit}, doesn't outperform cuSPARSE~\cite{naumov2010cusparse}. MergePath~\cite{shan2023mergepath} addresses SpMM workload imbalance with a binary search algorithm, but its efficacy decreases with feature dimensions over 128, common in large graphs. Flow-GNN~\cite{sarkar2022flowgnn} accelerates GNNs on FPGA platforms but fails to address workload imbalance or provide scalability to larger core count.

Several methods have been developed to tackle large graph problems, such as graph neighborhood and boundary sampling. Betty~\cite{yang2023betty} offers a novel sampler to alleviate memory bottlenecks, and BNS-GCN~\cite{wan2022bns} uses boundary sampling for multi-GPU and multi-node systems; yet both fail to address the SpMM bottleneck. Other research has utilized generalized SpMM for GNN inference acceleration, including AWB-GCN~\cite{geng2020awb}, which dynamically balances workload, and I-GCN~\cite{geng2021gcn}, which enhances locality and reduces off-chip memory access. However, the implementation requires specialized hardware and is not applicable to GPU system.

\subsection{Introducing Sparsity in GNN Training}

\textbf{Dropout:}
As a well-known regularization to prevent overfitting, dropout introduces feature sparsity during training by randomly setting a fraction of input units to 0~\cite{baldi2013understanding}. This operation results in a form of sparsity that is highly irregular and challenging to leverage in an system/hardware design.

\textbf{Weight Sparsification:}
Two prevailing weight sparsification approaches in GNN training are \textit{train-and-prune} and \textit{sparse training}~\cite{peng2022towards}. The former optimizes weight parameters to improve inference speed, as exemplified by methods like ADMM-based pruning~\cite{zhang2018systematic} and LTH-based pruning~\cite{frankle2018lottery}. The overall training cost (including pretraining) is usually much higher compared to the original model training. Conversely, \textit{sparse training} initiates with a sparsified weight matrix and updates sparse weight locations at specific iterations. The standard scheduler, {\em drop and grow}, includes techniques like SET~\cite{mocanu2018scalable}, RigL~\cite{evci2020rigging}, and SNFS~\cite{dettmers2019sparse}. Such sparsification in GNN workload, however, introduces irregular patterns that inhibit efficient hardware deployment~\cite{chen2021unified}.

\textbf{Nonlinearity for Sparsification:}
Nonlinear functions such as ReLU~\cite{agarap2018deep} introduce sparsity into the graph training. As detailed in FATReLU~\cite{kurtz2020inducing}, adjusting the ReLU threshold can induce greater feature sparsity. Similar to other sparsity forms, this irregularity does not align with hardware characteristics, yielding limited speedup on the training system.

\section{\ourframework Dataflow}

We start with introducing MaxK nonlinearity, and how it can benefit GNN training system in \ourframework dataflow.

\subsection{MaxK Nonlinearity as a Universal Approximator} 
Conventional ReLU operators, which are frequently utilized in GNN architectures, result in an irregularly sparsified feature matrix, thereby hindering its usage for hardware acceleration.  To address this challenge, we introduce the MaxK nonlinearity.

\textbf{MaxK Nonlinearity Definition:}
\textit{(i) During the forward propagation, MaxK nonlinearity is computed on node-wise feature map to get the maximum $k_{th}$ element and set the rest to 0. (ii) During the backward propagation, the feature gradient uses same feature sparsity pattern as induced in forward. }

{\small
\begin{equation}
h(X) = \textnormal{\textit{max-k}}_{j \in [1,r]} (X \cdot W + b)_{j}
 \label{eq:maxk}
\end{equation}
}
where $W$, $b$, and $X$ represent weights, biases, and inputs, representatively. $h(X)$ denotes intermediate feature, which is a piece-wise linear (PWL) function of $X$. $r$ is the hidden dimension/number of neurons.
In addition, the MaxK nonlinear operator is positioned before the SpMM operator, which serves to diminish the computational and memory access overhead associated with SpMM. This nonlinearity also exhibits generalization ability in both transductive and inductive graph learning settings.  It introduces a regularized sparsity pattern, enabling a more efficient design for hardware acceleration.

\begin{figure}[t!]
\centering
    \includegraphics[
    clip, 
    width =0.96 \linewidth]{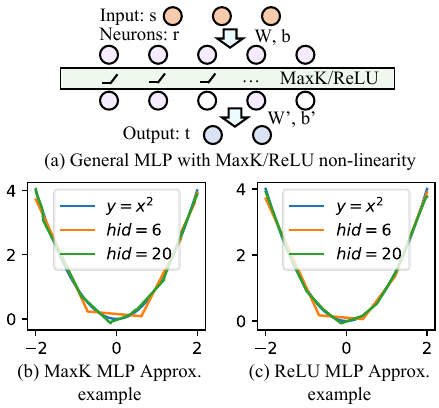} 
     \caption{
     \revA{
     MLP with MaxK and ReLU non-linearity. $y = x^2$ function approximation example with different number of hidden units.} }
    \label{fig:maxk_nn} 
\vspace{-4mm}
\end{figure}

MaxK, like ReLU, could be represented using a PWL function. We use Multilayer Perceptron (MLP) with MaxK nonlinearity to
theoretically analyze the universal approximation characteristics of MaxK nonlinearity.
As demonstrated in Eq.~\ref{eq:maxk} and as illustrated in Fig.~\ref{fig:maxk_nn}(a), MaxK is applied to the feature map. $X$ has a size $s$. $W$ has dimensions $[s, r]$. MaxK maintains the maximum $k$ significant value out of $r$ while preserving the same shape. 
MaxK preserves the same input and output dimensions and introduces a regularized sparsity pattern, thereby facilitating the design of the supporting hardware. MaxK can approximate any continuous function $f(X)$ of $X \in \mathbb{R}^s$ with a sufficient number of hidden units $r$.

\begin{proposition}
Given any positive integers $s$ (input dimension) and $t$ (output dimension), two parameter groups $W$ and $W'$ are determined such that $g(X)$ is expressed as a linear combination of $r$ convex PWL functions:
\small
\begin{equation}
g(X) = h(X) \cdot W' + b'
\end{equation}
$g(X)$ operates as the neural network approximator and denotes the continuous PWL function with $r$ locally affine regions on $\mathbb{R}^s$. Thus,  $g(X)$ (could be any continuous PWL function) can be represented as a linear combination of $h(X)$ (proof can be found in \cite{wang2004general, goodfellow2013maxout}).
\label{prop1}
\end{proposition}

\begin{theorem} \emph{MLP with MaxK Serves as a Universal Approximator}. 
A MaxK network $g(X)$ with $r$ hidden units can approximate any continuous function $f(X)$ on a compact domain $C\subset \mathbb{R}^s$ with an arbitrarily small approximation error $\epsilon$. In particular, as $\epsilon \rightarrow 0$, it follows that $r \rightarrow \infty$.
\end{theorem}

\begin{figure*}[ht]
\centering
    \includegraphics[width =0.94 \linewidth]{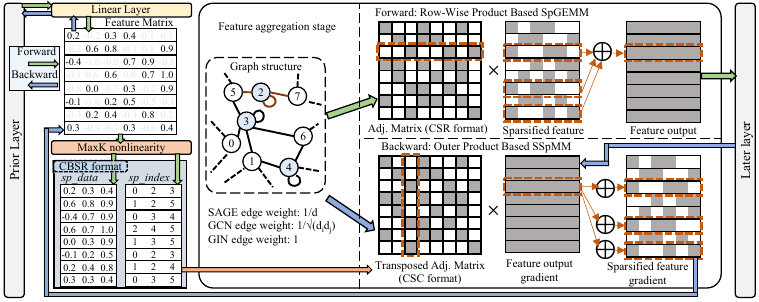} 
     \caption{
     Training dataflow of single MaxK based GNN layer. In the backward computation, the transposed CSC format is equal to original CSR format. } 
    \label{fig:forward_backward} 
\vspace{-2mm}
\end{figure*}

\textbf{Universal Approximator}: 
Based on the Stone-Weierstrass approximation theorem~\cite{de1959stone, goodfellow2013maxout}, a PWL function $g(X)$ can approximate any continuous function $f(X)$ with an error $\epsilon$: $\left| f (v) - g(v) \right| < \epsilon$. The PWL function $g(X)$, as provided in Proposition \ref{prop1}, is composed of $r$ number of
PWL functions $h(X)$. By configuring a sufficiently large number of hidden units $r$ and appropriate $k$ values in the MaxK network $g(X)$, the desired approximation error $\epsilon$ can be achieved. Consequently,
a MaxK network with $r$ hidden units could provide an arbitrarily close approximation to $f(X)$ on the compact domain $C\subset \mathbb{R}^s$.

Refer to Fig.~\ref{fig:maxk_nn}(b) for an illustration of a single-layer MaxK-based network employed for the approximation of the function $y = x^2$. A standard backpropagation algorithm is used to train the neural network until convergence, with the number of hidden units varied to observe the approximation error. For the MaxK nonlinearity, the top $\lceil hid/4 \rceil$ elements are selected and the remainder set to 0. It can be readily observed that as the number of hidden units increases, the approximation error of the MaxK-based neural network approximator decreases. Fig.~\ref{fig:maxk_nn}(c) shows the $y = x^2$ function approximation using ReLU nonlinearity, results show that ReLU and MaxK nonlinearity have a similar approximation performance.

\takeaway{The \ourframework system introduces the regularized sparsity of the embedding feature matrix, thereby considerably speeding up GNN models' SpMM operation. Notably, MaxK is a nonlinearity and does not compromise the precision of the models significantly.}

{
\subsection{\ourframework Training Dataflow} 
Traditional GNN layers adopt similar backward and forward SpMM computation designs. However, our proposed \ourframework framework incorporates asymmetric forward and backward paths, leveraging MaxK nonlinearity. 
MaxK nonlinear operator is positioned before the SpMM operator, sparsifying the forward computation path from SpMM to SpGEMM. Similarly, MaxK nonlinearity also increases sparsity of the backward computation path to SSpMM. Such a process significantly reduces the computational and memory overhead associated with matrix multiplications.
To leverage this newly introduced embedding sparsity, we present modified forward and backward propagation dataflow and generate a reference kernel design tailored for \ourframework. Fig.~\ref{fig:forward_backward} illustrates the dataflow of single-layer GCN training with the proposed MaxK nonlinearity, for the sake of simplicity.

Given a GNN layer with single linear and aggregation operations,
the forward and backward processes of the aggregation stage are expressed in Eq.~\ref{eq:backward_spmm}, where $A$ is the adjacent list and $h(X_{l-1})$ denotes the feature map post linear layer and the MaxK nonlinearity. $A$ could have different expressions according to aggregator type. For instance, the SAGEConv uses $1/d$ ($d$ is the node degree) for the mean aggregator. The backward process inherents the sparsity pattern as we only compute the gradient of $h(X_{l-1})$ non-zero elements.

{\small
\begin{equation}\label{eq:backward_spmm}
 X_{l} = A \cdot h(X_{l-1}) , \:\:\:  \frac{\partial L}{\partial h(X_{l-1})} = A^T \cdot \frac{\partial L}{\partial X_{l}} 
\end{equation}
}

\textbf{Forward Computation}. For the forward computation, we propose employing a Compressed Balanced Sparse Row (CBSR) format to capitalize on the newly introduced sparsity of the feature output resulting from the MaxK layer. The CBSR format allows for contiguous memory accesses on sparsified feature matrix and improves the system memory bandwidth utilization while mitigating workload imbalance. This format comprises two components: a data segment ($sp\_data$) and an index segment ($sp\_index$), which are stored in two adjacent memory blocks in the main memory. The next step involves the execution of forward feature aggregation, accomplished by multiplying the graph adjacency list by the sparsified feature matrix. This computation utilizes a row-wise product-based SpGEMM scheme~\cite{srivastava2020matraptor}, whereby $X_{l}[i, :] = \sum_{j=0}^{J}A[i, j]\cdot h(X_{l-1})[j, :]$. Assuming a dense output obviates the costly \revC{ESC overhead}~\cite{dalton2015optimizing} usually encountered with SpGEMM design. During the row-wise product operation, each element from the left-hand row is multiplied by its corresponding elements in the right-hand row, with the result then accumulated to the output row. This procedure enables the sparsified output accumulation to occur within the on-chip cache, offering significantly lower latency compared to global memory-based accumulation.

\textbf{Backward Computation}. During the feature aggregation's backward SSpMM process, the transposed adjacency matrix, in a CSC format (equivalent to CSR employed in forward), is multiplied with the feature output gradient to yield the sparsified feature gradient. Compared with standard SpGEMM computations, this backward SSpMM computation exhibits a (sparse $\times$ dense = sparse) operation.
Given that the output sparsity pattern ($sp\_index$) aligns with that of the forward process, the backward SpGEMM only requires to compute corresponding data ($sp\_data$) located by ($sp\_index$).

However, the row-wise product-based multiplication for this computation could lead to substantial irregular global memory access on 
$\frac{\partial L}{\partial X_{l}}$ (from Eq.~\ref{eq:backward_spmm}), as elements must be fetched according to the sparse index $sp\_index$. To mitigate this, we propose prefetching rows from $\frac{\partial L}{\partial X_{l}}$ to the on-chip memory, enabling irregular access within the cache, thereby avoiding uncoalesced global memory traffic.

Consequently, we adopt an outer product-based~\cite{srivastava2020matraptor} SSpMM method for the backward computation process, where
\begin{equation}
    \frac{\partial L}{\partial h(X_{l-1})}[:, :] = \sum_{j=0}^{J}A^T[:, j]\cdot \frac{\partial L}{\partial X_{l}} [j, :]
\end{equation}
Each element of the left-hand column is multiplied by a single row and accumulated to the respective output row. This strategy ensures efficient utilization of memory and alleviates the irregular memory access issue.
}

{
\section{GPU System Support for \ourframework}

\subsection{Forward SpGEMM GPU Kernel}

Compared to conventional SpMM, our proposed CBSR format incorporates a sparsified input embedding matrix, with the aim of reducing both computational and memory demands. Sparse input matrices correlate to an increase in irregular memory accesses to both the input matrices (adjacency matrix and embedding matrix) which could degrade kernel computational, memory efficiency. To counteract the lack of spatial locality in the CSR formatted adjacency list and the CBSR formatted embedding matrix, we use a \emph{warp-level partitioning scheme}. Coupled with an on-chip buffering mechanism, this approach can achieve warp-level balance and coalesced global memory accesses, significantly improving computational efficiency.

\begin{figure}[t] 
    \includegraphics[
    clip,
    width =0.98 \linewidth]{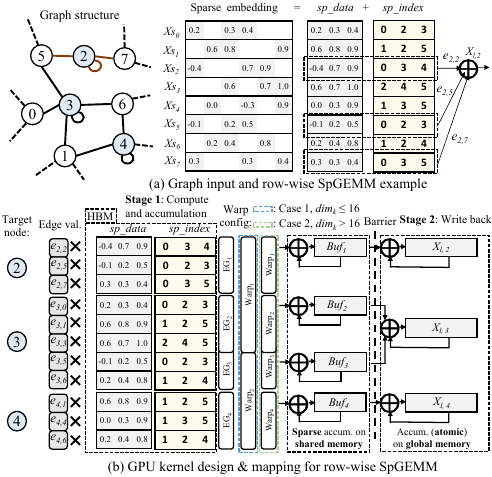}
    \vspace{-2mm}
     \caption{
     Forward computation kernel with $dim_{origin}=6$ and $dim_k=3$. $Xs_i$: ith row (node) of input embedding $Xs$, represented in the CBSR format. $X_{l,i}$: ith row (node) of output embedding $X_l$.}
    \label{fig:forward_kernel} 
    \vspace{-3mm}
\end{figure}

\textbf{Design Overview}. Fig.~\ref{fig:forward_kernel}(a) shows an illustration of the row-wise SpGEMM computation with the CBSR format.
The computation of node 2's output embedding is based on the multiplication and accumulation of neighboring embeddings ($Xs_2$, $Xs_5$, $Xs_7$) and corresponding adjacency list edge values ($e_{2, 2}$, $e_{2, 5}$, $e_{2, 7}$). The accumulation leverages the index $sp\_index$ to map multiplication results to the appropriate output positions, which consequently results in a sparse memory access pattern. Therefore, we buffer the partial accumulation result in the on-chip shared memory to mitigate uncoalesced global memory transactions.

The on-chip buffering considerations are encapsulated in the kernel design pseudocode provided in Algorithm~\ref{alg:row_spmm}. The overview workflow is divided into two stages: i) a compute and accumulation stage, and ii) a write back stage. Suppose the original dimensions of the right matrix $X_s$ are denoted as $dim_{origin}$ and the MaxK value selected results in $dim_k$, leading to a CBSR formatted matrix. This matrix has $sp\_data$ and $sp\_index$, each with a size of $N \times dim_k$. Within the computation and accumulation kernel, the coalesced global memory transactions fetch both $sp\_data$ and $sp\_index$. The parallel multiplication and sparsified accumulation within the warp are conducted within $Buf_w$, which locates in the shared memory. Eventually, $Buf_w$ is added into $X_{l}$ in global memory using coalesced accesses. This design strategy promotes an efficient memory access pattern, optimizing computational parallelism while conserving memory bandwidth.

{
\begin{algorithm}[t!]
\caption{Forward Computation Kernel Pseudocode}
\label{alg:row_spmm}
\begin{algorithmic}[1]
\small
\FOR{all rows $A_{row\_i}$ in $A$}
 \FOR{all warp partitions $P_w$ in $A_{row\_i}$}
  \STATE Initialize $Buf_w$ in shared memory;
  \STATE Form $m$ threads within a warp; 
  \FOR{each nonzero element $e_{i,j}$ in $P_w$}
   \FOR{all $thread_k$ in warp ($k \in [1, m]$)} 
    \STATE \mycomment{// Multiply and sparse accumulation to $Buf_w$, mapped by $sp\_index$} 
    \STATE $Buf_w[sp\_index[j, k]] \pluseq e_{i,j} \times sp\_data[j, k]$;
   \ENDFOR
  \ENDFOR
 \ENDFOR
 \STATE Reorganize all threads by natural warps;
 \FOR{all $Buf_w$ in shared memory}
 \STATE \mycomment{// Atomically accumulation with coalesced global memory access} 
 \STATE $X_{l,i} \pluseq Buf_{w}$;
 \ENDFOR
\ENDFOR
\end{algorithmic}
\end{algorithm}
}

\textbf{Warp Level Partition}.
Illustrated in Algorithm~\ref{alg:row_spmm}, the SpGEMM workload requires a workload-to-warp mapping strategy. Herein, we delve into the warp-level workload partitioning and allocation.
We propose a light-weight warp-level partition mapper that operates at $\mathcal{O}(n)$ complexity with $n$ being the number of nodes.

As shown in Fig.~\ref{fig:forward_kernel}(b), each edge $e_{i, j}$ involved in the computation constitutes a workload unit. Within such a unit, $e_{i, j}$ undergoes a multiplication with the sparse row $sp\_data_{j}$, followed by accumulation in the buffer $Buf_w$, indexed by $sp\_index_{j}$. Subsequently, the workload of each adjacency matrix row $A_{row_i}$ is segmented into Edge Groups ($EGs$). Each $EG$ reserves a chunk of shared of ($dim_{origin} \times 4$) bytes to serve as an intermediate buffer for sparse accumulation.

The workload of each adjacency matrix row $A_{row_i}$ is firstly segmented into Edge Groups (EGs). 
To optimize EG execution within the computation and accumulation phase, we integrate the hidden dimension into our warp mapping strategy. In scenarios where $dim_k \leq 16$ (as seen in Case 1 of Fig.\ref{fig:forward_kernel}(b)), each standard warp comprises $\lfloor \frac{32}{dim_k} \rfloor$ EG workloads, EG is limited to be within the same warp to circumvent memory access conflicts that could occur if an EG straddles multiple warps. 
In contrast, if $dim_k > 16$ (Case 2 in Fig.\ref{fig:forward_kernel}(b)), an EG is processed by a single warp executed iteratively. The execution of each EG is performed by the corresponding warp, with the results aggregated at respective locations in the shared memory buffer, following the indices from $sp\_index$.

In the next phase (stage 2 in Fig.~\ref{fig:forward_kernel}), data from each shared memory buffer $Buf_w$ is atomically accumulated into its corresponding $X_l$ output in global memory. Retaining the thread organization of natural warps due to identical dimensions of the shared memory buffers and output embeddings ($dim_{origin}$), each warp cyclically processes a single row, ensuring an efficient and coalesced computational structure.
}

\begin{figure}[t!] 
    \includegraphics[
    clip,
    width =0.98 \linewidth]{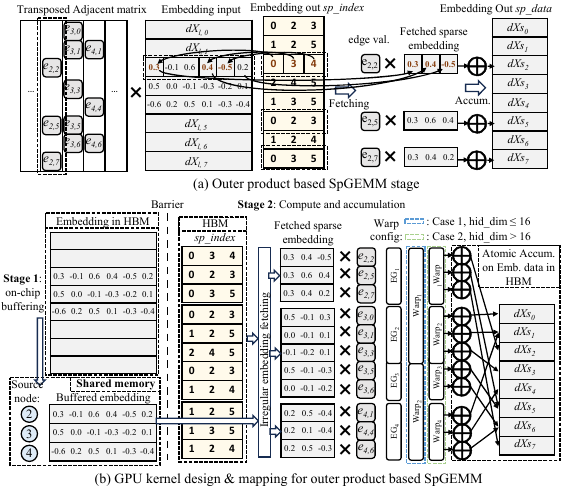} 
    \vspace{-2mm}
     \caption{
     Backward computation kernel with $dim_{origin}=6$ and $dim_k=3$. $dX_{l,i}$: ith row (node) of dense input embedding $dX_{l}$. $dXs_{i}$: ith row (node) of output embedding $dXs$ represented in the CBSR format. Transposed adjacent matrix $A^T$ in the CSC format has same storage format as the original adjacent matrix $A$ in CSR format, thus no extra storage. }
    \label{fig:backward_kernel} 
    \vspace{-3mm}
\end{figure}

{
\subsection{Backward SSpMM GPU Kernel}

In the backward computation, we execute a specialized SSpMM operation involving $A^T$ $\times$ $dX_l$ to generate $dXs$, in CBSR format. Inheriting $sp\_index$ from $Xs$ used in the forward computation, the computation needs only $sp\_data$ of $dXs$, signifying a (sparse $\times$ dense = sparse) operation with a known output sparse pattern, thereby requiring only data locations indexed by $sp\_index$. A naive row-wise product-based kernel could lead to significant uncoalesced global memory transactions, thereby inhibiting data parallelism.

\textbf{Design Overview}. We propose utilizing an outer product-based SSpMM process to enhance global memory coalescing and computational parallelism. An illustrative dataflow is presented in Fig.~\ref{fig:backward_kernel}(a). Here, the third column of $A^T$, denoted by ($e_{2, 2}$, $e_{2, 5}$, $e_{2, 7}$), multiplies with the third input node embedding, $dX_{l,2}$, indexed by $sp\_index$ (represented as $sp\_index_{row\_2}$, $sp\_index_{row\_5}$, and $sp\_index_{row\_7}$). The resulting values are subsequently accumulated into the corresponding output embedding data $sp\_data$ (expressed as $dXs_{2}$, $dXs_{5}$, $dXs_{7}$). The transposed adjacency matrix, $A^T$, is represented in CSC format, mirroring the data structure of the original matrix $A$ in CSR format. Consequently, this approach requires no additional memory for storing the backward gradient computation, thereby optimizing memory utilization. Considering the irregular indexing produced by $sp\_index$, buffering the dense embedding row $dX_{l}$ in on-chip memory proves advantageous, as it enhances bandwidth and reduces latency, due to more regular memory access.

\begin{algorithm}
\caption{Backward Computation Kernel Pseudocode}
\label{alg:outer_spmm}
\begin{algorithmic}[1]
\small
\FOR{all rows $dX_{l, row\_i}$ in $dX_l$}
 \FOR{all workload partitions $P_w$ in $A^T_{col\_i}$}
  \STATE \mycomment{// Coalesced global memory read.}
  \STATE Load $dX_{l, row\_i}$ into $Buf_w$ in shared memory;
  \STATE Form $m$ threads within a warp
  \FOR{each nonzero element $e_{i,j}$ in $P_w$}
   \FOR{all $thread_k$ ($k \in [1, m]$) in warp} 
    \STATE \mycomment{// 
Collect data from the buffer $Buf_w$, indexed by $sp\_index$, and multiply it by $e_{i, j}$. Subsequently, perform an automatic accumulation to $sp\_data$, ensuring coalesced global memory access.
    }
    \STATE $sp\_data[i, k] \pluseq e_{i,j} \times Buf_w[sp\_index[i, k]]$
   \ENDFOR
  \ENDFOR
 \ENDFOR
\ENDFOR
\end{algorithmic}
\end{algorithm}

\begin{table*}[htbp]
\centering
\caption{Graph datasets number of nodes and number of edges information. }
\resizebox{0.99\textwidth}{!}{
\begin{tabular}{|c|c|c|c|c|c|c|c|c|c|c|c|}
\hline
Graph Name & \# Nodes  & \# Edges   & Graph Name      & \# Nodes  & \# Edges   & Graph Name      & \# Nodes  & \# Edges & Graph Name & \# Nodes  & \# Edges    \\ \hline
am         & 881,680   & 5,668,682  & amazon0505      & 410,236   & 4,878,874  & amazon0601      & 403,394   & 5,478,357   & artist     & 50,515    & 1,638,396   \\ \hline
citation   & 2,927,963 & 30,387,995 & collab          & 235,868   & 2,358,104  & com-amazon      & 334,863   & 1,851,744   & DD         & 334,925   & 1,686,092  \\ \hline
ddi        & 4,267     & 2,135,822  & Flickr          & 89,250    & 989,006    & ogbn-arxiv      & 169,343   & 1,166,243   & ogbn-products & 2,449,029 & 123,718,280 \\ \hline
ogbn-proteins & 132,534 & 79,122,504 & OVCAR-8H       & 1,889,542 & 3,946,402  & ppa             & 576,289   & 42,463,862  & PROTEINS\_full & 43,466   & 162,088 \\ \hline
pubmed     & 19,717    & 99,203     & ppi             & 56,944    & 818,716    & Reddit      & 232,965   & 114,615,891 & SW-620H      & 1,888,584 & 3,944,206 \\ \hline
TWITTER-Partial & 580,768 & 1,435,116 & Yeast          & 1,710,902 & 3,636,546  & Yelp            & 716,847   & 13,954,819  & youtube       & 1,138,499 & 5,980,886  \\ \hline
\end{tabular}}
\label{tab:graph_details_revised}
\end{table*}

Our proposed design for backward computation kernel, presented in Algorithm~\ref{alg:outer_spmm}, encompasses two primary stages. The \textbf{first stage} involves loading the dense embedding $dX_{l, row\_i}$ into the shared memory buffer $Buf_w$ for each warp. It is crucial to note that this stage facilitates coalesced and continuous global memory transactions, optimizing memory access patterns. The \textbf{secondary stage} amalgamates sparse fetching, computation, and atomic accumulation. Sparse fetching encompasses two operations: a) fetching $sp\_index$ through coalesced global memory transactions, and b) irregular indexing within the shared memory buffer $Buf_w$. The fetched vector subsequently undergoes multiplication with the corresponding edge values $e_{i, j}$. Finally, the result is atomically accumulated in the global memory embedding data section $sp\_data$, a coalesced memory transaction that ensures computational efficiency.

\textbf{Warp Level Partitioning}. 
To enhance computational efficiency and ensure warp-level workload balance, we advocate for an edge-centric grouping process, analogous to the scheme used for the forward SpGEMM. 
IT is a lightweight process that can be seamlessly applied during the graph loading and preprocessing stage, preserving overall resource efficiency.

During the dense embedding loading stage, each warp fetches the corresponding row of the embedding matrix, $dX_{l, row\_i}$, into shared memory through coalesced global memory access. The required shared memory allocation per warp for floating-point numbers is $dim_{origin} \times 4$ bytes, mirroring the allocation used in the forward workflow. To expedite loading, we utilize a natural warp organization, with each warp iteratively managing its corresponding row. Afterward, all warps arrive at a synchronization barrier.

During the compute and accumulation stage, workload units working with edges and the hidden dimension are reconfigured into EGs, adopting the same partitioning procedure used in the forward computational workflow. In cases where $dim_k \leq 16$ (refer to Case 1 in Fig.~\ref{fig:backward_kernel}(b)), each conventional warp manages $\lfloor \frac{32}{dim_k} \rfloor$ EGs, confining each EG to a single warp to prevent shared memory access conflicts. For $dim_k > 16$ (Case 2 in Fig.~\ref{fig:backward_kernel}(b)), each EG is handled by a single warp using a loop function. Operations involving sparse fetching, computation, and atomic accumulation are performed within these warps, as outlined in Algorithm~\ref{alg:outer_spmm}. This stage exclusively involves coalesced memory read/write operations, thus preserving the efficiency of global memory transactions.

}

\subsection{Memory System}
\label{sec:memory}

In our design, the NVIDIA GPU's shared memory is strategically utilized to mitigate uncoalesced memory accesses and to ensure that all global memory accesses are coalesced. The memory system is structured to store the CSR-formatted adjacent matrix, the embedding matrix, and the CBSR-formatted sparse embedding matrix in global memory (HBM), while the intermediate shared memory serves as a buffer for partial results and sparse fetching. In this section, we examine the global memory transactions for both the forward SpGEMM and the backward SSpMM kernels.

\textbf{Forward SpGEMM}. 
During the forward SpGEMM computation, the bulk of the computation and sparse fetching is focused on the accumulation process within the shared memory. By implementing a row-wise product-based SpGEMM kernel design, the CBSR-formatted $X_s$ rows are read $nnz$ times, leading to a total global memory traffic of $(4 \times 2 \times dim_{k} \times nnz)$ bytes for floating-point data and integer index. With smaller $dim_{origin}$, utilizing uint8 for $sp\_index$ allows a reduction in total traffic to $(5 \times dim_{k} \times nnz)$ bytes. 
Compared to a row-wise SpMM kernel design, the total \textbf{global memory traffic reduction} is calculated as $[(4 \times dim_{origin} - 5 \times dim_{k}) \times nnz]$ bytes, indicating that lower values of $dim_{k}$ yield greater reductions.

\begin{figure*}[t]
\centering
    \includegraphics[width =0.98 \linewidth]{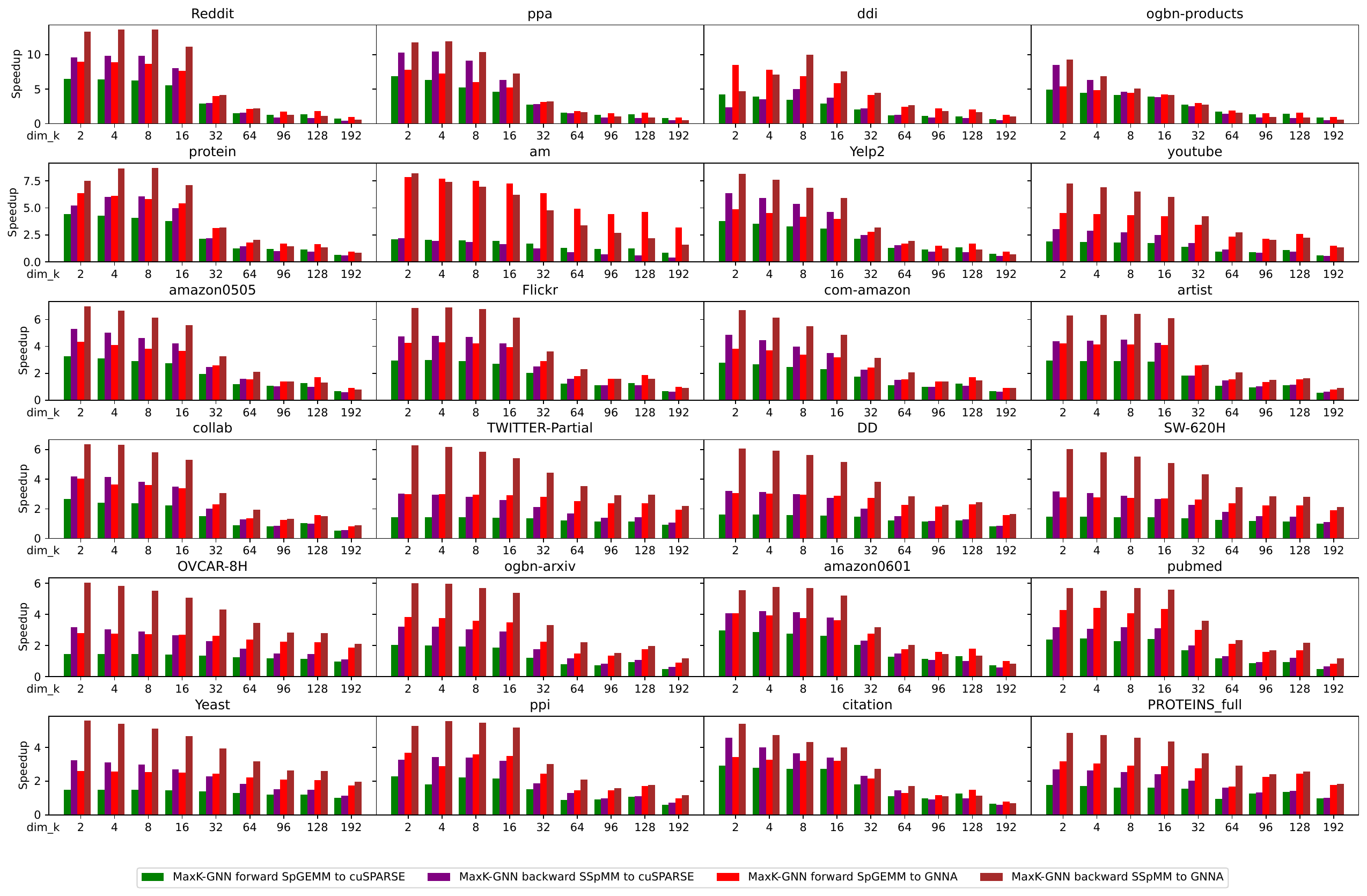} 
    \vspace{-3mm}
     \caption{Forward SpGEMM and backward SSpMM speedup over SPMM kernel from cuSPARSE~\cite{naumov2010cusparse} and GNNAdvisor~\cite{wang2021gnnadvisor}. \revA{Original hidden dimension size is 256, we vary $dim_k$ ($k$ value of MaxK) to evaluate kernel speedup. }
     }
    \label{fig:kernel_speedup} 
\end{figure*}

Additionally, the output atomic accumulation in our proposed SpGEMM kernel aligns with the original row-wise SpMM kernel, where the number of global memory atomic accumulations is given by $(N \times dim_{origin} \times \frac{avgdeg}{w})$, and $avgdeg$ is derived as $\frac{nnz}{N}$ with $w$ representing the hyperparameter for the maximum workload units assigned to an EG.

\textbf{Backward SSpMM}. 
The backward SSpMM begins with an on-chip buffering stage, allowing the buffered feature gradient row $dX_{l,i}$ to be read only once per SSpMM computation, equivalent to $(N \times dim_{origin})$ memory transactions. Subsequent stages require reading the corresponding rows from $sp\_index$ for sparse fetching, and during the compute and accumulation stages, each workload unit performs single read and write operations for its corresponding row in $sp\_data$. Consequently, the total read and write transactions to global memory are approximately $(4 \times N \times dim_{origin} + 5 \times dim_k \times nnz)$ and $(4 \times dim_k \times nnz)$ bytes respectively, when considering uint8 $sp\_index$. Compared to a naive outer product-based SpMM, the \textbf{global memory traffic reduction} is $[(4 \times dim_{origin} - 5 \times dim_k) \times nnz]$ for reads and $[(4 \times dim_{origin} - 4 \times dim_k) \times nnz]$ for write transactions, reaffirming that a lower $dim_{k}$ leads to higher reductions.

\subsection{Kernel Profiling}
\revA{
To demonstrate the effectiveness of the proposed design, We provide profiling of SpGEMM and SSpMM kernels, while the experiment setup is outlined in Sec.~}\ref{sec:eval}.
\revA{For similicity, we use Reddit graph as an example and provide the memory system profiling result shown in Table}~\ref{tab:memory_profile}. \revA{We evaluate the compute kernels by employing the Nsight Compute profiler to generate performance metrics for cuSPARSE SpMM, SpGEMM, and SSpMM kernels when executed on the Reddit graph. Table}~\ref{tab:memory_profile}\revA{ reports data on the traffic between the L2 cache and global memory, as well as the hit rates for the L1/L2 caches.}

\begin{table}[t!]
\centering
\small
\caption{\ourframework \revA{memory system profiling}}
\resizebox{0.95\columnwidth}{!}{
\begin{tabular}{c|c|c|c}
\hline
\begin{tabular}[c]{@{}c@{}}dim\_org = 256\\ dim\_k = 32\end{tabular}                           & SpMM      & SpGEMM    & SSpMM     \\ \hline
Total Traffic (GB)                                                                             & 138.05    & 13.13     & 14.02     \\ \hline
L1 cache hit rate (\%)                                                                         & 1.53      & 22.16     & 28.27     \\ \hline
L2 cache hit rate (\%)                                                                         & 51.75     & 75.44     & 89.43     \\ \hline
\begin{tabular}[c]{@{}c@{}}Memory bandwidth \\ utilization (\%) \end{tabular} & 60.90 & 33.60 & 48.08 \\ \hline
\end{tabular}
}
\label{tab:memory_profile}
\end{table}

\revA{
The reported memory traffic reduction of the proposed SpGEMM and SSpMM kernel aligns with the theoretical analysis given in Section}~\ref{sec:memory}\revA{. The proposed MaxK nonlinearity and corresponding kernel support reduces total global memory traffic by close to 90.5\% / 89.8\%, when reducing original hidden dimension from 256 to $k$ as 32. While the traffic is reduced significantly, the bandwidth utilization is not reduced significantly, as such we are able to achieve \textbf{2.9x/2.98x} speedup over cuSPARSE SpMM.}

\revA{
It is worth mentioning that the L1/L2 cache hit rates of cuSPARSE SpMM kernel, our forward SpGEMM kernel, and our backward SSpMM kernel are 1.53\%/51.75\%, 22.16\%/75.44\%, and 28.27\%/89.43\%, respectively. The L1 cache hit rate of our kernels is significantly higher than that of cuSPARSE SpMM kernel, which is due to our rational use of the multi-level memory hierarchy of the GPU.
}

\section{Evaluation}
\label{sec:eval}
We offer a comprehensive assessment of \ourframework, highlighting its performance advantages and applicability in the broader context of graph-based learning and computation.
Our evaluation strategy begins with an in-depth analysis of both the forward
SpGEMM
and the backward 
SSpMM kernels. We carefully assess these components under a range of $K$ values, comparing the performance with existing implementations such as SpMM, as found in GNNAdvisor~\cite{wang2021gnnadvisor} and cuSPARSE v12.0~\cite{naumov2010cusparse}.
Following this detailed kernel-level examination, we present our system  training time evaluation. This encapsulates both the accuracy and speedup metrics to showcase the efficacy of the \ourframework framework in addressing general graph learning problems.

\begin{figure*}[t]
\centering
    \includegraphics[width =0.98 \linewidth]{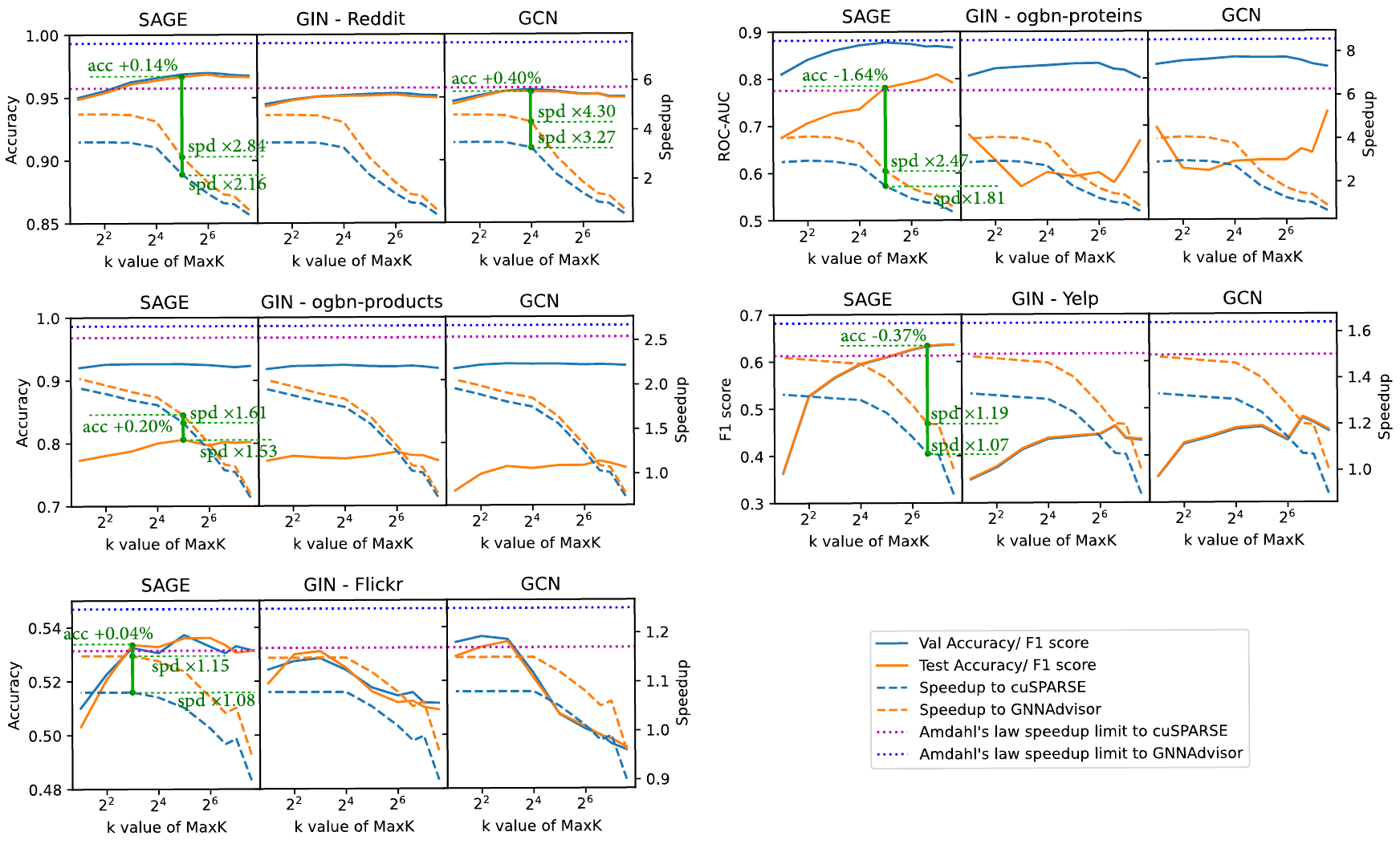} 
     \vspace{-2mm}
     \caption{
      \ourframework system evaluation for GraphSAGE~\cite{hamilton2017inductive}, GCN~\cite{kipf2016semi}, and GIN~\cite{xu2018gin} models and Reddit, ogbn-proteins, ogbn-produces, Yelp, and Flickr datasets. MaxK nonlinearity setting: $k = [2, 4, 8, 16, 32, 64, 96, 128, 192]$. We compare the \ourframework training speedup over DGL with cuSPARSE~\cite{dgl_amazon} framework and GNNAdvisor~\cite{wang2021gnnadvisor}. spd: speedup. 
     }
    \label{fig:acc_speedup} 
    \vspace{-1mm}
\end{figure*}

\subsection{Experimental Setup}

\textbf{Datasets}.
For SpGEMM \& SSpMM kernel benchmark, we select popular benchmark datasets that have been extensively employed in previous studies~\cite{hamilton2017inductive, kipf2017semi, xu2019how, kersting2016benchmark, fey2019fast, leskovec2014snap, wang2021gnnadvisor}. See Table~\ref{tab:graph_details_revised}.

For \ourframework system evaluation, we benchmark five datasets that range from small to medium-scale graphs. 
Specifically, the selected datasets are Flickr~\cite{JulianFlickr}, for the categorization of image types based on descriptors and shared attributes; Yelp~\cite{Zengiclr20}, for the classification of user-generated reviews pertaining to businesses and services; Reddit~\cite{reddit}, for community prediction using posts' content and users' comments; ogbn-products~\cite{hu2020open}, for Amazon product classification via customer reviews; and ogbn-proteins~\cite{hu2020open}, for the prediction of protein function presence. This suite of datasets not only covers a broad spectrum of applications but also facilitates the understanding of how \ourframework performs under various scenarios.

\textbf{Models}.
For SpGEMM \& SSpMM kernel benchmarks, we employ an original hidden dimension of 256. This evaluation involves an assessment of the MaxK sparsified feature matrix across a set of $k$ values, including $[2, 4, 8, 16, 32, 64, 96, 128, 192]$. The selected parameters ensure a comprehensive examination of the impact of various sparsity levels on the system's performance.

\begin{table}[ht!]
\centering
\small
\caption{\ourframework training setup}
\vspace{-3mm}
\resizebox{0.99\columnwidth}{!}{
\begin{tabular}{c|c|c|c|cc}
\hline
\multirow{2}{*}{Dataset} & \multirow{2}{*}{Flickr} & \multirow{2}{*}{Yelp} & \multirow{2}{*}{Reddit} & \multicolumn{2}{c}{OGB}                   \\ \cline{5-6} 
                         &                         &                       &                         & \multicolumn{1}{c|}{products}  & proteins \\ \hline
Layers                   & 3                       & 4                     & 4                       & \multicolumn{1}{c|}{3}         & 3        \\ \hline
Hid. dim.              & 256                     & 384                   & 256                     & \multicolumn{1}{c|}{256}       & 256      \\ \hline
Epochs                   & 400                     & 3000                  & 3000                    & \multicolumn{1}{c|}{500}       & 1000     \\ \hline
LR/Drop               & 0.001/0.2               & 0.001/0.1             & 0.01/0.5                & \multicolumn{1}{c|}{0.003/0.5} & 0.01/0.5 \\ \hline
\end{tabular}
}
\label{tab:learning_setup}
\vspace{-1mm}
\end{table}

For the system  training time evaluation of \ourframework, we integrate MaxK nonlinearity with three graph models: GraphSAGE~\cite{hamilton2017inductive}, GCN~\cite{kipf2016semi}, and GIN~\cite{xu2018gin}. The GIN model~\cite{xu2018gin} is noteworthy for its unique aggregation function, serving as a reference for advanced GNNs such as Graph Attention Networks (GAT)~\cite{GATConv}. To ensure fairness, all models are trained in full batch graph learning mode with the MEAN aggregator for GraphSAGE. The detailed parameter settings of \ourframework are given in Table~\ref{tab:learning_setup}. Our evaluation spans $k = [2, 4, 8, 16, 32, 64, 96, 128, 192]$ to find the best trade-off between model accuracy and system speedup, highlighting the versatility of our approach.

\textbf{Environment Setup}. We construct the \ourframework by implementing the SpGEMM \& SSpMM kernels using C++ and CUDA C. Our principal evaluation platform consists of a high-performance server equipped with 32-core AMD EPYC 7513 CPU and NVIDIA A100 80GB GPU~\cite{choquette2021nvidia}.
This computing environment is utilized for both kernel-level and system-level evaluations.

For the performance measurement of the SpGEMM \& SSpMM kernels, we conduct a rigorous analysis by calculating the average latency across \textit{\textbf{1000 runs}}. This ensures that the observed performance metrics are consistent and representative. In the context of the comprehensive evaluation of the \ourframework system, our approach is further extended. We measure the latency of the training phase with epochs given in Table~\ref{tab:learning_setup}, including both forward and backward propagation, 50 times, and obtain the average runtime.

\subsection{\ourframework Kernel Evaluation}

We presente forward SpGEMM and backward SSpMM kernel evaluation in Fig.~\ref{fig:kernel_speedup}. Both SpGEMM and SSpMM kernels exhibit a significant speedup compared to the SpMM kernels from cuSPARSE~\cite{naumov2010cusparse} and GNNAdvisor~\cite{wang2021gnnadvisor}. Note that the original hidden dimension is 256 and we vary $k$ values for the benchmark.
Overall, the result shows that as $k$ decreases, the speedup increases. 

For the SpGEMM kernel, when $k$ is decreased to a certain extent, such as 8, the output accumulation stage becomes the performance bottleneck, hence a further decrease in $k$ leads to a speedup saturation. Nevertheless, our SpGEMM kernel exhibits impressive acceleration performance, especially on graphs with larger average degrees. For graphs with an average degree greater than 50, such as ogbn-proteins, ddi, Reddit, ppa, and ogbn-products, the average speedup of the SpGEMM kernel at $k=8, 16, 32, 64$ is $4.63\times$, $4.15\times$, $2.54\times$, $1.46\times$, respectively, as compared to the cuSPARSE~\cite{naumov2010cusparse}, and $6.39\times$, $5.71\times$, $3.50\times$, $2.02\times$, respectively, as compared to the GNNAdvisor~\cite{wang2021gnnadvisor}, demonstrating its high acceleration efficiency. When $k\leq128$, the SpGEMM kernel can effectively bring speedup to 92.2\% of all test cases compared to cuSPARSE~\cite{naumov2010cusparse}, and 100\% of all test cases compared to GNNAdvisor~\cite{wang2021gnnadvisor}. The result demonstrates its high generalization.


The backward SSpMM kernel fully exploits the global memory access coalescing and traffic reduction by combining the outer product-based approach with dense row prefetching. The SSpMM kernel achieves better speedup performance than the forward SpGEMM kernel. For graphs with average degrees greater than 50, the average speedup of the SSpMM kernel at $k=8, 16, 32, 64$ is $6.93\times$, $5.39\times$, $2.55\times$, $1.46\times$ respectively, as compared to the cuSPARSE~\cite{naumov2010cusparse} and $9.57\times$, $7.46\times$, $3.55\times$, $2.04\times$, respectively, as compared to the GNNAdvisor~\cite{wang2021gnnadvisor}. Given that a \ourframework layer training pipeline requires computing both forward SpGEMM and the backward SSpMM, the SSpMM kernel speedup can also benefit the training speedup. When $k\leq128$, the SSpMM kernel can effectively bring speedup to 87.5\% of all test cases compared to cuSPARSE~\cite{naumov2010cusparse}, and 98.4\% of all test cases compared to GNNAdvisor~\cite{wang2021gnnadvisor}. It also exhibits significant versatility.

\subsection{ \ourframework System  Training Time Evaluation}

\revC{\textbf{MaxK Nonlinearity Kernel}. For MaxK nonlinearity kernel implementation, we customize a high-performance pivot-based $k_{th}$ value selection kernel. The kernel buffers each node’s embedding in shared memory and finds the $min$ and $max$ values 
Then the kernel selects a pivot equal to $(min + max)/2$ and counts the number of elements that are greater than the pivot. The algorithm iterates based on the pivot value until the number of elements (that are greater than the pivot) is equal to $k$. The feature map distribution is not fully randomized, and follows a normal distribution. With 256 original hidden dimensions, we observe that the pivot based algorithm usually converges when running feature map MaxK selection in less than 10 iterations. Given that the comparison operations and pivot selection are conducted in shared memory, the total global memory traffic would be similar to element-wise operations such as ReLU. The overall MaxK nonlinearity kernel has an average cost of less than 2\% of the SpGEMM kernel run time. We provide an example of kernel run-time on the Reddit within Table}~\ref{tab:maxk_latency}.

\begin{table}[ht!]
\centering
\small
\caption{\revC{MaxK nonlinearity kernel profiling}}
\resizebox{0.95\columnwidth}{!}{
\begin{tabular}{c|c|c|c|l}
\hline
\begin{tabular}[c]{@{}c@{}}dim\_org = 256\\ dim\_k = 32\end{tabular} & SpMM  & SpGEMM & SSpMM & MaxK  \\ \hline
Latency (ms)                                                         & 44.98 & 15.49  & 15.07 & 0.261 \\ \hline
\end{tabular}
}
\label{tab:maxk_latency}
\end{table}

\revC{
Morever, the MaxK nonlinearity kernel is applied during forward path, and the sparse matrix index can be shared with backward propagation process. We need to “recompress” feature into CBSR format for each GNN layer during the forward path. However, the MaxK nonlinearity kernel has little overhead compared to SpMM and SpGEMM/SSpMM, and it will not become the critical path during the training pipeline. 
}

\textbf{Accuracy and Speedup}. To comprehensively evaluate the \ourframework framework, we conduct experiments using three representative GNN models: GraphSAGE\cite{hamilton2017inductive}, GCN\cite{kipf2016semi}, and GIN\cite{xu2018gin}. 
We use five diverse datasets to benchmark performance. 
\revD{The ReLU-based \textbf{baseline} model's setting and accuracy utilized in our evaluation section is aligned with the SOTA full-batch training accuracy. Specifically, our baseline performance matches the results presented in Table 3 of reference}~\cite{liu2023rsc}\revD{ and the GraphSAGE row in Table 4 of reference}~\cite{wan2022bns}.
We test \ourframework system with $k = [2, 4, 8, 16, 32, \\ 64, 96, 128, 192]$, as shown in Figure~\ref{fig:acc_speedup}. In the figure, we provide speedup limit lines calculated using Amdahl's law~\cite{gustafson1988reevaluating}: $S = 1/(1 - p\_{SpMM})$, where $S$ is the speedup limit and $p\_{SpMM}$ represents the percentage of execution time taken by the SpMM operator within the full GNN training pipeline. This allows us to contextualize the empirical speedups achieved by \ourframework.

\begin{table*}[htbp]
\centering
\caption{\centering \ourframework training accuracy \& speedup evaluation and comparison with ReLU based baseline model implemented in DGL~\cite{dgl_amazon}}
\resizebox{0.99\textwidth}{!}{
\begin{tabular}{cc|ccc|ccc|ccc|ccc|ccc}
\hline
\multicolumn{2}{c|}{dataset}                          & \multicolumn{3}{c|}{Reddit}                                                                                                                                                               & \multicolumn{3}{c|}{ogbn-proteins}                                                                                                            & \multicolumn{3}{c|}{ogbn-products}                                                                                                                                                        & \multicolumn{3}{c|}{Yelp}                                                                                                                       & \multicolumn{3}{c}{Flickr}                                                                                                                                                                \\ \hline
\multicolumn{1}{c|}{model}                 & method   & \multicolumn{1}{c|}{k}  & \multicolumn{1}{c|}{\begin{tabular}[c]{@{}c@{}}Acc\\ (\%)\end{tabular}} & \begin{tabular}[c]{@{}c@{}}Latency\\ (ms/epoch)\\ Speedup\\ (cuSP./GNNA.)\end{tabular} & \multicolumn{1}{c|}{k}  & \multicolumn{1}{c|}{AUC}    & \begin{tabular}[c]{@{}c@{}}Latency\\ (ms/epoch)\\ Speedup\\ (cuSP./GNNA.)\end{tabular} & \multicolumn{1}{c|}{k}  & \multicolumn{1}{c|}{\begin{tabular}[c]{@{}c@{}}Acc\\ (\%)\end{tabular}} & \begin{tabular}[c]{@{}c@{}}Latency\\ (ms/epoch)\\ Speedup\\ (cuSP./GNNA.)\end{tabular} & \multicolumn{1}{c|}{k}  & \multicolumn{1}{c|}{F1 score} & \begin{tabular}[c]{@{}c@{}}Latency\\ (ms/epoch)\\ Speedup\\ (cuSP./GNNA.)\end{tabular} & \multicolumn{1}{c|}{k}  & \multicolumn{1}{c|}{\begin{tabular}[c]{@{}c@{}}Acc\\ (\%)\end{tabular}} & \begin{tabular}[c]{@{}c@{}}Latency\\ (ms/epoch)\\ Speedup\\ (cuSP./GNNA.)\end{tabular} \\ \hline
\multicolumn{1}{c|}{\multirow{3}{*}{SAGE}} & baseline & \multicolumn{1}{c|}{-}  & \multicolumn{1}{c|}{96.51}                                              & \begin{tabular}[c]{@{}c@{}}54.9 \\ (1x/1.32x)\end{tabular}                            & \multicolumn{1}{c|}{-}  & \multicolumn{1}{c|}{0.7976} & \begin{tabular}[c]{@{}c@{}}23.4\\ (1$\times$/1.37$\times$)\end{tabular}               & \multicolumn{1}{c|}{-}  & \multicolumn{1}{c|}{80.39}                                              & \begin{tabular}[c]{@{}c@{}}133.6\\ (1$\times$/1.05$\times$)\end{tabular}              & \multicolumn{1}{c|}{-}  & \multicolumn{1}{c|}{0.6376}   & \begin{tabular}[c]{@{}c@{}}36.5\\      (1$\times$/1.11$\times$)\end{tabular}          & \multicolumn{1}{c|}{-}  & \multicolumn{1}{c|}{53.31}                                              & \begin{tabular}[c]{@{}c@{}}3.52\\ (1$\times$/1.06$\times$)\end{tabular}               \\ \cline{2-17} 
\multicolumn{1}{c|}{}                      & \textbf{MaxK-GNN}  & \multicolumn{1}{c|}{32} & \multicolumn{1}{c|}{96.65}                                              & \begin{tabular}[c]{@{}c@{}}25.5\\ (2.16$\times$/2.84$\times$)\end{tabular}            & \multicolumn{1}{c|}{64} & \multicolumn{1}{c|}{0.7928} & \begin{tabular}[c]{@{}c@{}}18.6\\ (1.25$\times$/1.71$\times$)\end{tabular}            & \multicolumn{1}{c|}{32} & \multicolumn{1}{c|}{80.59}                                              & \begin{tabular}[c]{@{}c@{}}87.1\\ (1.53$\times$/1.61$\times$)\end{tabular}            & \multicolumn{1}{c|}{96} & \multicolumn{1}{c|}{0.6339}   & \begin{tabular}[c]{@{}c@{}}34.2\\      (1.07$\times$/1.19$\times$)\end{tabular}       & \multicolumn{1}{c|}{32} & \multicolumn{1}{c|}{53.6}                                               & \begin{tabular}[c]{@{}c@{}}3.35\\ (1.05$\times$/1.12$\times$)\end{tabular}            \\ \cline{2-17} 
\multicolumn{1}{c|}{}                      & \textbf{MaxK-GNN}  & \multicolumn{1}{c|}{16} & \multicolumn{1}{c|}{96.37}                                              & \begin{tabular}[c]{@{}c@{}}17\\ (3.22$\times$/4.24$\times$)\end{tabular}              & \multicolumn{1}{c|}{32} & \multicolumn{1}{c|}{0.7812} & \begin{tabular}[c]{@{}c@{}}12.9\\ (1.81$\times$/2.47$\times$)\end{tabular}            & \multicolumn{1}{c|}{16} & \multicolumn{1}{c|}{80}                                                 & \begin{tabular}[c]{@{}c@{}}77.9\\ (1.72$\times$/1.80$\times$)\end{tabular}            & \multicolumn{1}{c|}{32} & \multicolumn{1}{c|}{0.61}     & \begin{tabular}[c]{@{}c@{}}29.6\\      (1.23$\times$/1.37$\times$)\end{tabular}       & \multicolumn{1}{c|}{8}  & \multicolumn{1}{c|}{53.35}                                              & \begin{tabular}[c]{@{}c@{}}3.26\\ (1.08$\times$/1.15$\times$)\end{tabular}            \\ \hline
\multicolumn{1}{c|}{\multirow{3}{*}{GCN}}  & baseline & \multicolumn{1}{c|}{-}  & \multicolumn{1}{c|}{95.02}                                              & \begin{tabular}[c]{@{}c@{}}54.5\\ (1$\times$/1.32$\times$)\end{tabular}               & \multicolumn{1}{c|}{-}  & \multicolumn{1}{c|}{0.646}  & \begin{tabular}[c]{@{}c@{}}23.2\\ (1$\times$/1.37$\times$)\end{tabular}               & \multicolumn{1}{c|}{-}  & \multicolumn{1}{c|}{76.58}                                              & \begin{tabular}[c]{@{}c@{}}129.6\\ (1$\times$/1.05$\times$)\end{tabular}              & \multicolumn{1}{c|}{-}  & \multicolumn{1}{c|}{0.4718}   & \begin{tabular}[c]{@{}c@{}}34.3\\ (1$\times$/1.12$\times$)\end{tabular}               & \multicolumn{1}{c|}{-}  & \multicolumn{1}{c|}{49.78}                                              & \begin{tabular}[c]{@{}c@{}}3.42\\ (1$\times$/1.06$\times$)\end{tabular}               \\ \cline{2-17} 
\multicolumn{1}{c|}{}                      & \textbf{MaxK-GNN}  & \multicolumn{1}{c|}{16} & \multicolumn{1}{c|}{95.42}                                              & \begin{tabular}[c]{@{}c@{}}16.7\\ (3.27$\times$/4.30$\times$)\end{tabular}            & \multicolumn{1}{c|}{16} & \multicolumn{1}{c|}{0.6236} & \begin{tabular}[c]{@{}c@{}}8.43\\ (2.75$\times$/3.77$\times$)\end{tabular}            & \multicolumn{1}{c|}{32} & \multicolumn{1}{c|}{76.34}                                              & \begin{tabular}[c]{@{}c@{}}83.2\\ (1.56$\times$/1.64$\times$)\end{tabular}            & \multicolumn{1}{c|}{96} & \multicolumn{1}{c|}{0.4819}   & \begin{tabular}[c]{@{}c@{}}32.0\\ (1.07$\times$/1.20$\times$)\end{tabular}            & \multicolumn{1}{c|}{8}  & \multicolumn{1}{c|}{53.45}                                              & \begin{tabular}[c]{@{}c@{}}3.17\\ (1.08$\times$/1.15$\times$)\end{tabular}            \\ \cline{2-17} 
\multicolumn{1}{c|}{}                      & \textbf{MaxK-GNN}  & \multicolumn{1}{c|}{8}  & \multicolumn{1}{c|}{95.46}                                              & \begin{tabular}[c]{@{}c@{}}15.7\\ (3.48$\times$/4.58$\times$)\end{tabular}            & \multicolumn{1}{c|}{2}  & \multicolumn{1}{c|}{0.6958} & \begin{tabular}[c]{@{}c@{}}7.94\\ (2.92$\times$/4.00$\times$)\end{tabular}            & \multicolumn{1}{c|}{8}  & \multicolumn{1}{c|}{76.21}                                              & \begin{tabular}[c]{@{}c@{}}71.6\\ (1.81$\times$/1.91$\times$)\end{tabular}            & \multicolumn{1}{c|}{32} & \multicolumn{1}{c|}{0.4628}   & \begin{tabular}[c]{@{}c@{}}27.5\\ (1.25$\times$/1.40$\times$)\end{tabular}            & \multicolumn{1}{c|}{4}  & \multicolumn{1}{c|}{53.25}                                              & \begin{tabular}[c]{@{}c@{}}3.16\\ (1.08$\times$/1.15$\times$)\end{tabular}            \\ \hline
\multicolumn{1}{c|}{\multirow{3}{*}{GIN}}  & baseline & \multicolumn{1}{c|}{-}  & \multicolumn{1}{c|}{95.07}                                              & \begin{tabular}[c]{@{}c@{}}54.6\\ (1$\times$/1.32$\times$)\end{tabular}               & \multicolumn{1}{c|}{-}  & \multicolumn{1}{c|}{0.583}  & \begin{tabular}[c]{@{}c@{}}23.3\\ (1$\times$/1.37$\times$)\end{tabular}               & \multicolumn{1}{c|}{-}  & \multicolumn{1}{c|}{77.79}                                              & \begin{tabular}[c]{@{}c@{}}130.7\\ (1$\times$/1.05$\times$)\end{tabular}              & \multicolumn{1}{c|}{-}  & \multicolumn{1}{c|}{0.4578}   & \begin{tabular}[c]{@{}c@{}}34.9\\ (1$\times$/1.12$\times$)\end{tabular}               & \multicolumn{1}{c|}{-}  & \multicolumn{1}{c|}{50.78}                                              & \begin{tabular}[c]{@{}c@{}}3.35\\ (1$\times$/1.07$\times$)\end{tabular}               \\ \cline{2-17} 
\multicolumn{1}{c|}{}                      & \textbf{MaxK-GNN}  & \multicolumn{1}{c|}{16} & \multicolumn{1}{c|}{95.11}                                              & \begin{tabular}[c]{@{}c@{}}16.7\\ (3.27$\times$/4.32$\times$)\end{tabular}            & \multicolumn{1}{c|}{4}  & \multicolumn{1}{c|}{0.6277} & \begin{tabular}[c]{@{}c@{}}7.81\\ (2.98$\times$/4.07$\times$)\end{tabular}            & \multicolumn{1}{c|}{8}  & \multicolumn{1}{c|}{77.69}                                              & \begin{tabular}[c]{@{}c@{}}72.7\\ (1.80$\times$/1.89$\times$)\end{tabular}            & \multicolumn{1}{c|}{96} & \multicolumn{1}{c|}{0.464}    & \begin{tabular}[c]{@{}c@{}}32.7\\ (1.07$\times$/1.20$\times$)\end{tabular}            & \multicolumn{1}{c|}{8}  & \multicolumn{1}{c|}{53.11}                                              & \begin{tabular}[c]{@{}c@{}}3.10\\ (1.08$\times$/1.15$\times$)\end{tabular}            \\ \cline{2-17} 
\multicolumn{1}{c|}{}                      & \textbf{MaxK-GNN}  & \multicolumn{1}{c|}{8}  & \multicolumn{1}{c|}{95.05}                                              & \begin{tabular}[c]{@{}c@{}}15.8\\ (3.47$\times$/4.57$\times$)\end{tabular}            & \multicolumn{1}{c|}{2}  & \multicolumn{1}{c|}{0.6812} & \begin{tabular}[c]{@{}c@{}}7.97\\ (2.92$\times$/3.99$\times$)\end{tabular}            & \multicolumn{1}{c|}{4}  & \multicolumn{1}{c|}{77.98}                                              & \begin{tabular}[c]{@{}c@{}}69.8\\ (1.87$\times$/1.97$\times$)\end{tabular}            & \multicolumn{1}{c|}{32} & \multicolumn{1}{c|}{0.4422}   & \begin{tabular}[c]{@{}c@{}}28.1\\ (1.24$\times$/1.39$\times$)\end{tabular}            & \multicolumn{1}{c|}{4}  & \multicolumn{1}{c|}{52.99}                                              & \begin{tabular}[c]{@{}c@{}}3.09\\ (1.08$\times$/1.15$\times$)\end{tabular}            \\ \hline
\end{tabular}
}
\label{tab:final_acc}
\end{table*}

Reddit and ogbn-proteins allow greater speedup due to their characteristics. Using a lower $k$ value with these datasets leads to a slight accuracy decline but permits substantial system speedup exceeding 3x with a suitable $k$ value selection. The ogbn-produces, Yelp, and Flickr datasets have relatively lower speedup limits. For these datasets, \ourframework achieves 1.1-2$\times$  speedup without significant accuracy loss. Lowering $k$ values trades off some accuracy for larger speedups on datasets with higher speedup limits like Reddit and ogbn-proteins. However, even on datasets with lower speedup limits, \ourframework provides 1.1-2$\times$ speedups with minimal accuracy impact.

We select the best performing $k$ values from \ourframework framework, aiming to further investigate the relationship between accuracy and system speedup. We compare the results against a ReLU-based baseline GNN model, which has been implemented in the DGL framework~\cite{dgl_amazon}. The results are encapsulated in Table~\ref{tab:final_acc}. 

GraphSAGE (SAGE) on Reddit has a speedup limit of 5.52$\times$/7.27$\times$ compared to cuSP./GNNA. respectively, following Amdahl's law~\cite{gustafson1988reevaluating}. We utilize \ourframework with $k=32$ and attain speedup factors of 2.16$\times$/2.84$\times$, resulting in enhanced accuracy for the GraphSAGE  model. In cases where \ourframework is implemented with $k=16$, speedup factors of 3.48$\times$/4.58$\times$ are achieved with the GCN model setting, while simultaneously elevating the accuracy by 0.44\%. GraphSAGE (SAGE) on Yelp possesses a lower Amdahl's law speedup limit, 1.46$\times$/1.59$\times$, compared to cuSP./GNNA. respectively. The dataset requires a relatively higher $k$ value to uphold accuracy performance. With the original hidden dimension of the Yelp dataset being 384, \ourframework with GraphSAGE \& $k = 96$ achieves 1.07$\times$/1.19$\times$ speedup relative to cuSP./GNNA. baselines, while maintaining comparable accuracy.
The Flickr dataset also manifests a lower Amdahl's law speedup limit, which is 1.16$\times$/1.24$\times$ compared to cuSP./GNNA. respectively. However, \ourframework with GraphSAGE \&
$k = 8$ acquires a 1.08$\times$/1.15$\times$ speedup, accompanied by greater accuracy.

The results collectively show that our \ourframework approaches the speedup limit. The performance gaps between our results and Amdahl's law theoretical limits, i.e., 3.22$\times$/4.24$\times$ compared to 5.52$\times$/7.27$\times$ for Reddit dataset using GraphSAGE, \revB{is from the essential accumulation stage of SpGEMM and dense row prefetching stage of SSpMM, which empirically are difficult to further optimize.}

\revD{\textbf{Further Discussion on Accuracy.} In Table}~\ref{tab:final_acc} \revD{and Figure}~\ref{fig:acc_speedup}\revD{, we follow the standard train/val/test split setting and obtain average accuracy over five random seeds for graph training. While most models and datasets demonstrate stable behavior, exceptions occur, notably with the ogbn-proteins dataset when using GCN/GIN models. Upon detailed examination, we attribute the inconsistent behavior observed in the ogbn-proteins dataset to inherent characteristics of the dataset itself. Specifically, within a certain range of the convergence region for the ogbn-proteins dataset, we observe high variance in test accuracy, which in turn leads to unstable ROC-AUC performance. Importantly, this observation is not exclusive to the MaxK-GNN model; we have also identified similar instability in the baseline models without MaxK.}

\revB{\textbf{$K$ Value Selection.} Empirically, we could select $k = 32$ for 256 original hidden dimension to align similar accuracy with ReLU baseline model and obtain significant kernel speedup. Such $K$ selection corresponds to 87.5\% feature sparsity.}

\begin{figure}[ht!]
\centering
    \includegraphics[
    clip, 
    width =0.85 \linewidth]{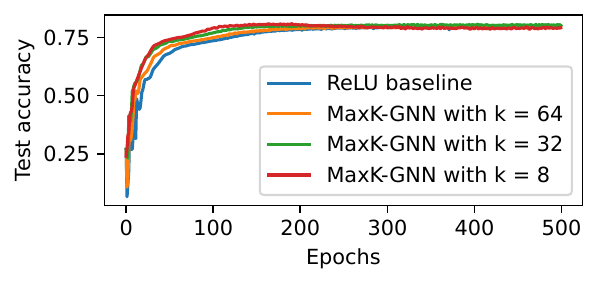} 
    \vspace{-3mm}
     \caption{
     Convergence curves of full-batch training on ogbn-product dataset for (i) ReLU baseline model (ii) \ourframework with $k = 64$ (iii)  \ourframework with $k = 32$ \revD{(iv) MaxK-GNN with $k = 8$} } 
    \label{fig:convergence} 
    \vspace{-1.2em}
\end{figure}

\textbf{Convergence Analysis of \ourframework}. To examine the convergence performance of the \ourframework training, we show a case study on the ogbn-products dataset with a full batch setting. The results in Fig.~\ref{fig:convergence} show that the \ourframework training, specifically at $k = 64$, $k = 32$ and $k = 8$, demonstrates convergence behavior similar or even better than the baseline model employing ReLU nonlinearity. With a lower $k$ value, the convergence speed is slightly faster.

\section{Conclusion and Future Work}

In this paper, we present \ourframework, a high-performance GPU training system integrating algorithm and system innovation. (i) We introduce the MaxK nonlinearity and provide a theoretical analysis of MaxK nonlinearity as a universal approximator, and present the Compressed Balanced Sparse Row (CBSR) format, designed to store the data and index of the feature matrix after nonlinearity; (ii) We design a coalescing enhanced forward computation
with row-wise product-based SpGEMM Kernel using CBSR for input feature matrix fetching and strategic placement of a sparse output accumulation buffer in shared memory; (iii) We develop an optimized backward computation with outer product-based and SSpMM Kernel.
Experiments show that our \ourframework system could approach the  limit according to Amdahl's law. We achieve comparable accuracy to existing GNNs, but at a significantly increased speed: 3.22$\times$/4.24$\times$ speedup (vs. theoretical limits, 5.52$\times$/7.27$\times$) on Reddit.

\revA{The proposed MaxK nonlinearity could be potentially expanded to more DNN architectures such as CNNs and Transformers, to provide regularly sparsified feature map for acceleration.}

\section{Acknowledgments}
This research was supported in part by the Northeastern University Institute for Experiential AI, the NSF IUCRC Center for Hardware and Embedded Systems Security and Trust (CHEST),
NSF SHF-2340273, the Semiconductor Research Corporation (SRC) Artificial Intelligence Hardware program, and Advanced Micro Devices (AMD).

%
%
%
%
%





\clearpage
\appendix
\section{Artifact Appendix}

\subsection{Abstract}

\ourframework has three main kernel design: 
The first part is the forward SpGEMM kernel design. The SpGEMM kernel reduces memory traffic via sparse on-chip accumulation thus lead to significant speedup. 
The second part is the backward SSpMM kernel design. We built outer-product based SSpMM kernel and use explicit dense row prefetching to reduce the overall memory traffic. 
The third part is the MaxK nonlinearity kernel design. We use pivot-based $k_{th}$ seeking algorithm in shared memory to build the kernel, and we iterate the algorithm for up to 10 iterations to find the $k_{th}$ split point.

\subsection{Artifact check-list (meta-information)}

{\small
\begin{itemize}
\itemsep0em 
    \item {\bf Algorithm: } MaxK nonlinearity design. Graph neural network (GNN) training.
    \item {\bf Program: } \ourframework for acceleration GNN training with MaxK nonlinearity. 
    \item {\bf Model: } GraphSage, GCN, GIN. 
    \item {\bf Data set: } Refer to Table~\ref{tab:graph_details_revised}.
    \item {\bf Run-time environment: } Ubuntu 20.04+
    \item {\bf Compilation: } GCC 9.4+, CMAKE 3.5+, CUDA 12.1+
    \item \textbf{Hardware}:
    \begin{itemize}
    \itemsep0em 
        \item CPU with x86\_64 architecture, host memory >= 256GB. Tested on 32-core AMD EPYC 7543 Processor (2-socket, total 64-core 128-thread) CPU with 2TB host memory.
        \item  NVIDIA GPU (arch>=$sm\_80$) with devcie memory >= 48GB. Tested on NVIDIA A100 ($sm\_80$). 
    \end{itemize}
    \item {\bf Execution: } Bash scripts
    \item {\bf Metrics: } Accuracy vs. latency under different MaxK $k$ value.
    \item {\bf Experiments: } Fig.~\ref{fig:kernel_speedup} and Fig.~\ref{fig:acc_speedup} in experiment section.  
    \item {\bf How much disk space required (approximately)?: } 128 GB
  \item {\bf How much time is needed to prepare workflow (approximately)?: } 2 hours
  \item {\bf How much time is needed to complete experiments (approximately)?: } 4 days
  \item {\bf Publicly available?: } Yes
  \item {\bf Code licenses (if publicly available)?: } MIT License
    \item {\bf Archived (DOI):} 10.5281/zenodo.10690770
\end{itemize}
}

\subsection{Description} 

\subsubsection{How to access} 

The artifact is available in \url{https://github.com/harveyp123/MaxK-GNN} and \url{https://zenodo.org/doi/10.5281/zenodo.10690770}

\subsubsection{Hardware dependencies}

\begin{itemize}
\itemsep0em 
    \item CPU with x86\_64 architecture, host memory >= 256GB. Tested on 32-core AMD EPYC 7543 Processor (2-socket, total 64-core 128-thread) CPU with 2TB host memory.
    \item  NVIDIA GPU (arch>=$sm\_80$) with devcie memory >= 48GB. Tested on NVIDIA A100 ($sm\_80$). 
\end{itemize}

\subsubsection{Software dependencies}
GCC 9.4+, CMAKE 3.5+, CUDA 12.1+, instruction to build conda environment can be found in github.

\subsection{Installation}

Download the source code: "\texttt{git clone \url{https://github.com/harveyp123/MaxK-GNN}}". Install environment according to \texttt{README.md} file.

\subsection{Experiment workflow}

For the SpGEMM, SSpMM and Maxk kernels benchmarking:

(1) Go to ./kernels directory and download the graph datasets. 

(2) Generate the meta-data for the SpGEMM and SSpMM kernels by executing generate\_meta.py . 

(3) Do the Compilation, then run the executable to benchmark the SpGEMM and SSpMM kernels. 

(4) Change the executable build configuration in CMakeLists.txt and re-do the compilation, then run the executable to benchmark the Maxk kernel.

The benchmarking results are outputted in console and can be logged to files using |tee command.
For more details, see \texttt{README.md} file.

Run training for \ourframework starting from \url{https://github.com/harveyp123/MaxK-GNN?tab=readme-ov-file#run-the-relu-baseline-training}. The experiment result will be logged in \texttt{./experiment} directory. For more details, see \texttt{README.md} file.

\subsection{Evaluation and expected results}

The code (\url{https://github.com/harveyp123/MaxK-GNN}) and instructions (\texttt{README.md}) can be used to reproduce Section~\ref{sec:eval} results for Fig.~\ref{fig:kernel_speedup} and Fig.~\ref{fig:acc_speedup}.



\bibliographystyle{unsrt}
\bibliography{refs, refs2}

\end{document}